\documentclass[10pt,twocolumn,letterpaper]{article}

\usepackage{cvpr}              

\usepackage{times}
\usepackage{epsfig}
\usepackage{graphicx}
\usepackage{amsmath}
\usepackage{amssymb}
\usepackage{bm}
\usepackage{booktabs}
\usepackage{multirow}
\usepackage{soul}
\usepackage[font=small]{caption}
\usepackage{subcaption}
\usepackage{enumitem}
\usepackage{mathtools}
\usepackage[accsupp]{axessibility}
\usepackage[ruled,vlined]{algorithm2e}

\newcommand{\Tref}[1]{Table~\ref{#1}}

\newlength\savewidth\newcommand\shline{\noalign{\global\savewidth\arrayrulewidth
  \global\arrayrulewidth 1pt}\hline\noalign{\global\arrayrulewidth\savewidth}}

\DeclareMathOperator*{\argmax}{arg\,max}

\usepackage[pagebackref,breaklinks,colorlinks]{hyperref}

\usepackage[capitalize]{cleveref}
\crefname{section}{Sec.}{Secs.}
\Crefname{section}{Section}{Sections}
\Crefname{table}{Table}{Tables}
\crefname{table}{Tab.}{Tabs.}

\def\cvprPaperID{5336} 
\def\confName{CVPR}
\def\confYear{2022}

\begin{document}

\title{Large-Scale Pre-training for Person Re-identification with Noisy Labels}

\author{Dengpan Fu$^1$ \quad
		Dongdong Chen$^3$ \quad
		Hao Yang$^2$ \quad
		Jianmin Bao$^2$\thanks{Corresponding author.} \\
		Lu Yuan$^3$ \quad
		Lei Zhang$^4$ \quad
		Houqiang Li$^1$ \quad
		Fang Wen $^2$ \quad
		Dong Chen$^2$
		\and 
		$^1$University of Science and Technology of China \quad $^2$Microsoft Research, $^3$Microsoft Cloud AI, $^4$IDEA  \\ 
		{\tt\small fdpan@mail.ustc.edu.cn} \quad {\tt\small cddlyf@gmail.com} \quad {\tt\small lihq@ustc.edu.cn} \\
		{\tt\small \{jianbao,haya,luyuan,fangwen,doch\}@microsoft.com, leizhang@idea.edu.cn}
	}

\maketitle

\begin{abstract}
This paper aims to address the problem of pre-training for person re-identification (Re-ID) with noisy labels. To setup the pre-training task, we apply a simple online multi-object tracking system on raw videos of an existing unlabeled Re-ID dataset ``LUPerson’’ and build the {\textit{\textbf{N}}}oisy {\textit{\textbf{L}}}abeled variant called ``LUPerson-{\textit{\textbf{NL}}}’’. Since theses ID labels automatically derived from tracklets inevitably contain noises, we develop a large-scale  {\textit{\textbf{P}}}re-training framework utilizing \textit{\textbf{N}}oisy \textit{\textbf{L}}abels ({\textit{\textbf{PNL}}}), which consists of three learning modules: supervised Re-ID learning, prototype-based contrastive learning, and label-guided contrastive learning. In principle, joint learning of these three modules not only clusters similar examples to one prototype, but also rectifies noisy labels based on the prototype assignment. We demonstrate that learning directly from raw videos is a promising alternative for pre-training, which utilizes spatial and temporal correlations as weak supervision. This simple pre-training task provides a scalable way to learn SOTA Re-ID representations from scratch on ``LUPerson-NL’’ without bells and whistles. For example, by applying on the same supervised Re-ID method MGN, our pre-trained model improves the mAP over the unsupervised pre-training counterpart by \textbf{5.7\%, 2.2\%, 2.3\%} on CUHK03,  DukeMTMC, and MSMT17 respectively. Under the small-scale or few-shot setting, the performance gain is even more significant, suggesting a better transferability of the learned representation. Code is available at \url{https://github.com/DengpanFu/LUPerson-NL}.
\end{abstract}

\vspace{-0.5cm}
\section{Introduction}
\label{sec:intro}

\begin{figure}[t]
\begin{center}
    \begin{subfigure}{1.62in}
        \centering\includegraphics[width=1.62in]{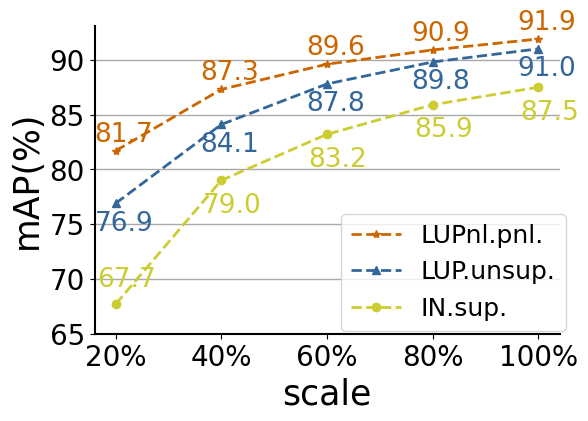}
        \vspace{-0.5cm}
        \caption{Market1501 with MGN}
    \end{subfigure}
    \begin{subfigure}{1.62in}
        \centering\includegraphics[width=1.62in]{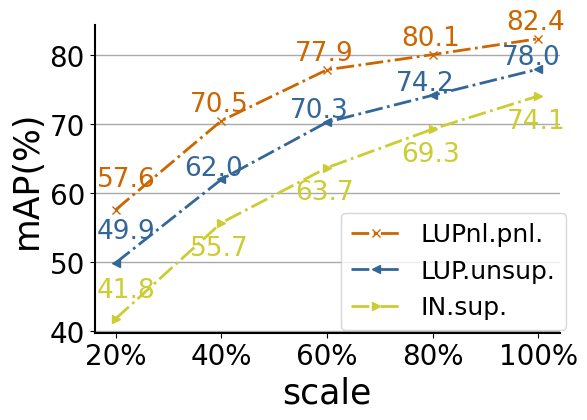}
        \vspace{-0.5cm}
        \caption{Market1501 with IDE}
    \end{subfigure}
    \begin{subfigure}{1.62in}
        \centering\includegraphics[width=1.62in]{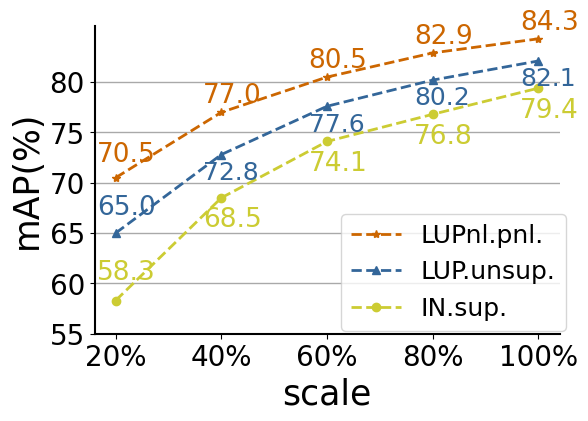}
        \vspace{-0.5cm}
        \caption{DukeMTMC with MGN}
    \end{subfigure}
    \begin{subfigure}{1.62in}
        \centering\includegraphics[width=1.62in]{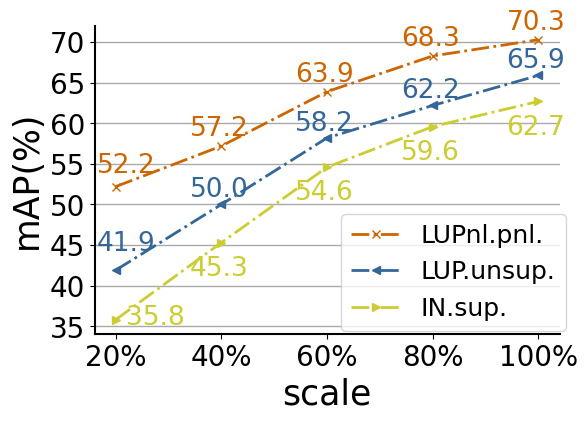}
        \vspace{-0.5cm}
        \caption{DukeMTMC with IDE}
    \end{subfigure}
\end{center}
\caption{
Comparing person Re-ID performances of three pre-trained models on two methods (IDE~\cite{zheng2017person} and MGN~\cite{wang2018learning}). Results are reported on Market1501 and DukeMTC, with different scales under the small-scale setting. \emph{IN.sup.} refers to the model supervised pre-trained on ImageNet, \emph{LUP.unsup.} is the model unsupervised pre-trained on LUPserson, and \emph{LUPnl.pnl.} is the model pre-trained on our LUPerson-NL dataset using our proposed PNL.
}
\label{fig:onecol}
\end{figure}

A large high-quality labeled dataset for person re-identification (Re-ID) is labor intensive and costly to create. Existing fully labeled datasets \cite{wei2018person,karanam2016comprehensive,Zheng2015ScalablePR,zheng2017unlabeled} for person Re-ID are all of limited scale and diversity compared to other vision tasks. Therefore, model pre-training becomes a crucial approach to achieve good Re-ID performance. However, due to the lack of large-scale Re-ID dataset, most previous methods simply use the models pre-trained on the crowd-labeled ImageNet dataset, resulting in a limited improvement because of the big domain gap between generic images in ImageNet and person-focused images desired by the Re-ID task. To mitigate this problem, the recent work \cite{fu2020unsupervised} has demonstrated that unsupervised pre-training on a web-scale unlabeled Re-ID image dataset ``LUPerson" (sub-sampled from massive streeview videos) surpasses that of pre-training on ImageNet.

In this paper, our hypothesis is that \emph{scalable ReID pre-training methods that learn directly from raw videos can generate better representations}. To verify it, we propose the \emph{noisy labels guided person Re-ID pre-training}, which leverages the spatial and temporal correlations in videos as weak supervision. This supervision is nearly cost-free, and can be achieved by the tracklets of a person over time derived from any multi-object tracking algorithm, such as~\cite{zhang2020fairmot}. In particular, we track each person in consecutive video frames, and automatically assign the tracked persons in the same tracklet to the same Re-ID label and vice versa. Enabled by the large amounts of raw videos in LUPerson~\cite{fu2020unsupervised}, publicly available data of this form on the internet, we create a new variant named {``LUPerson-{\bf{NL}}''} with derived pseudo Re-ID labels from tracklets for pre-training with noisy labels. This variant totally consists of $10M$ person images from $21K$ scenes with noisy labels of about $430K$ identities.

We demonstrate that contrastive pre-training of Re-ID is an effective method of learning from this weak supervision at large scale. This new \textbf{P}re-training framework utilizing \textbf{N}oisy \textbf{L}abels (\textbf{PNL}) composes three learning modules: (1) a simple \emph{supervised learning module} directly learns from Re-ID labels through classification; (2) a \emph{prototype-based contrastive learning module} helps cluster instances to the prototype which is dynamically updated by moving averaging the centroids of instance features, and progressively rectify the noisy labels based on the prototype assignment. and (3) a \emph{label-guided contrastive learning module} utilizes the rectified labels subsequently as the guidance. In contrast to the vanilla momentum contrastive learning \cite{he2020momentum,chen2020mocov2,fu2020unsupervised} that treats only features from the same instance as positive samples, our label-guided contrastive learning uses the rectified labels to distinguish positive and negative samples accordingly, leading to a better performance. In principle, joint learning of these three modules make the consistency between the prototype assignment from instances and the high confident (rectified) labels, as possible as it can.

The experiments show that our PNL model achieves remarkable improvements on various person Re-ID benchmarks. Figure~\ref{fig:onecol} indicates that the performance gain from our pre-trained models is consistent on different scales of training data. For example, upon the strong MGN~\cite{wang2018learning} baseline, our pre-trained model improves the mAP by $4.4\%, 4.9\%$ on Market1501 and DukeMTMC over the ImageNet supervised one, and $0.9\%,2.2\%$ over the unsupervised pre-training baseline \cite{fu2020unsupervised}. Moreover, the gains are even larger under the small-scale and few-shot settings, where the labeled Re-ID data are extremely limited. To the best of our knowledge, we are the first to show that large-scale noisy label guided pre-training can significantly benefit person Re-ID task.

Our key contributions can be summarized as follows:\vspace{-0.6em}
\begin{itemize}[leftmargin=*]
\setlength\itemsep{-0.2em}

\item We propose noisy label guided pre-training for person Re-ID, which incorporates supervised learning, prototype-based contrastive learning, label-guided contrastive learning and noisy label rectification to a unified framework.
\item We construct a large-scale noisy labeled person Re-ID dataset ``LUPerson-NL'' as a new variant of ``LUPerson". It is by far the largest noisy labeled person Re-ID dataset without any human labeling effort.
\item Our models pre-trained on LUPerson-NL push the state-of-the-art results on various public benchmarks to a new limit without bells and whistles. 
\end{itemize}

\section{Related Work}
\label{sec:related}
\noindent\textbf{Supervised Person Re-ID.} 
Most studies of person Re-ID employ supervised learning. Some \cite{hermans2017defense,chen2017beyond,yuan2020defense} introduce a hard triplet loss on the global feature, ensuring a closer feature distance for the same identity, while some \cite{zheng2017person,shen2018person,zheng2017discriminatively} impose classification loss to learn a global feature from the whole image. There are also some other works that learn part-based local features with separate classification losses. For example, Suh \etal \cite{suh2018part} presented part-aligned bi-linear representations and Sun \etal \cite{sun2018PCB} represented features as horizontal strips.
Recent approaches investigate learning invariant features concerning views \cite{liu2019view}, resolutions \cite{li2019recover}, poses \cite{li2019cross}, domains \cite{huang2019sbsgan,jin2020style}, or exploiting group-wise losses \cite{luo2019spectral} or temporal information \cite{gu2019temporal,li2019global} to improve performance.
The more advantageous results on public benchmarks are achieved by MGN \cite{wang2018learning}, which learns both global and local features with multiple losses. In \cite{qian2018pose}, Qian et al further demonstrated the potential of generating cross-view images for person re-indentification conditioned on normalized poses. In this paper, we focus on model pre-training, and our pre-trained models can be applied to these representative methods and boost their performance.

\noindent\textbf{Unsupervised Person Re-ID.}
To alleviate the lack of precise annotations, some works resort to unsupervised training on unlabeled datasets.
For example, MMCL \cite{wang2020unsupervised} formulates unsupervised person Re-ID as a multi-label classification to progressively seek true labels. BUC \cite{lin2019bottom} jointly optimizes the network and the sample relationship with a bottom-up hierarchical clustering.
MMT \cite{ge2019mutual} collaboratively trains two networks to refine both hard and soft pseudo labels.
SpCL \cite{ge2020selfpaced} designs a hybrid memory to unify the representations for clustering and instance-wise contrastive learning.
Both MMT \cite{ge2019mutual} and SpCL \cite{ge2020selfpaced} rely on explicit clustering of features from the whole training set, making them quite inefficient on large datasets like MSMT17.
Since the appearance ambiguity is difficult to address without supervision, these unsupervised methods have limited performance.
One alternative to address this issue is introducing model \emph{pre-training on large scale data}.
Inspired by the success of self-supervised representation learning \cite{wu2018unsupervised,he2020momentum,chen2020simple,chen2020big,chen2020mocov2,grill2020bootstrap,li2020mopro}, Fu \etal \cite{fu2020unsupervised} proposed a large scale unlabeled Re-ID dataset, LUPerson, and illustrated the effectiveness of its unsupervised pre-trained models.
In this work, we further try to make use of \emph{noisy labels} from video tracklets to improve the pre-training quality through large-scale weakly-supervised pre-training. 

\noindent\textbf{Weakly Supervised Person Re-ID.}
Several approaches also employ weak supervision in person Re-ID training.
Instead of requiring bounding boxes within each frame, Meng \etal \cite{meng2019weakly} rely on precise video-level labels, which reduces annotation cost but still need manual efforts to label videos.
On the contrary, we resort to noisy labels that can be automatically generated from tracklets on a much larger scale.
Some \cite{li2019unsupervised,wang2020weakly,chen2019weakly} also leverage tracklets to supervise the training of Re-ID tasks.
But unlike these approaches, we are proposing a \textbf{\emph{large-scale pre-training}} strategy for person Re-ID, by both building a new very large-scale dataset and devising a new pre-training framework:
the new dataset, LUPerson-NL, is even larger than LUPerson \cite{fu2020unsupervised} and has large amount of noisy Re-ID labels;
The new framework, PNL, combines supervised learning, label-guided contrastive learning and prototype based contrastive learning to exploit the knowledge under large-scale noise labels. 
Most importantly, our pre-trained models have demonstrated remarkable performance and generalization ability, 
helping achieve state-of-the-art results superior to all existing methods on public person Re-ID benchmarks. 

\vspace{-0.3em}
\section{LUPerson-NL: LUPerson With Noisy Labels}
\label{sec:lup-nl}
\vspace{-0.2em}

\begin{table*}[t]
    \centering
    \small
    \setlength\tabcolsep{4.5pt}
    \begin{tabular}{c|c|c|c|c|c|c|c|c}
    \shline
        Datasets & \#images & \#scene & \#persons & labeled & environment & camera view  & detector & crop size \ \\
        \hline
        VIPeR~\cite{gray2008viewpoint} & 1,264 & 2 & 632 & yes & - & fixed  & hand & $128\times48$ \\
        GRID~\cite{loy2013person} & 1,275 & 8 & 1,025 & yes & subway & fixed & hand & vary \\
        CUHK03~\cite{li2014deepreid} & $14,096$ & $2$ & $1,467$ & yes & campus & fixed & DPM\cite{felzenszwalb2009object}+hand & vary\\
        Market~\cite{Zheng2015ScalablePR} & $32,668$ & $6$ & $1,501$ & yes & campus & fixed & DPM\cite{felzenszwalb2009object}+hand & $128\times64$ \\
        Airport~\cite{karanam2016comprehensive} & $39,902$ & $6$ & $9,651$ & yes & airport & fixed & ACF\cite{dollar2014fast} & $128\times64$ \\
        DukeMTMC~\cite{zheng2017unlabeled} & $36,411$ & $8$ & $1,852$ & yes & campus & fixed & Hand &  vary \\
        MSMT17~\cite{wei2018person} & $126,441$ & $15$ & $4,101$ & yes & campus & fixed & FasterRCNN\cite{ren2015faster} & vary \\
        SYSU30K~\cite{wang2020weakly} & 29,606,918 & 1,000 & 30,508 & weakly & TV program & dynamic & YOLOv2 & vary \\
        LUPerson~\cite{fu2020unsupervised} & $4,180,243$ & $46,260$ & $>200k$ & no & vary & dynamic & YOLOv5 & vary \\
        \hline
        \textbf{LUPerson-NL} & $10,683,716$ & $21,697$ & $\simeq433,997$ & noisy & vary & dynamic & FairMOT\cite{zhang2020fairmot} & vary \\
        \shline
    \end{tabular}
    \caption{Comparing statistics among existing popular Re-ID datasets. LUPerson-NL is by far the largest Re-ID dataset with better diversity without human labeling effort. SYSU30K is partly annotated by human annotator.}
    \label{tab:data-stat}
\end{table*}

Supervised models based on deep networks are always data-hungry, but the labeled data they rely on are expensive to acquire.
It is a tremendous issue for person Re-ID task, since the human labelers need to check across multiple views to ensure the correctness of Re-ID labels.
The data shortage is partially alleviated by a recently published dataset, LUPerson \cite{fu2020unsupervised}, a dataset of unlabeled person images with a significantly larger scale than previous person Re-ID datasets. Unsupervised pre-trained models \cite{fu2020unsupervised} on LUPerson have demonstrated remarkable effectiveness without utilizing additional manual annotations, which arouses our curiosity: \emph{can we further improve the performance of pre-training directly by utilizing temporal correlation as weak supervision?} To verify this, we build a new variant of LUPerson on top of the raw videos from LUPerson and assign label to each person image with automatically generated tracklet. We name it \textbf{LUPerson-NL} with \textbf{NL} standing for \textbf{N}oisy \textbf{L}abels. It consists of $10M$ images with about $430K$ identities collected from $21K$ scenes. To our best knowledge, this is the largest person Re-ID dataset constructed without human labelling by far. Our \textbf{LUPerson-NL} will be released for scientific research only, while any usage for other purpose is forbidden. 

\subsection{Constructing LUPerson-NL}
\label{ssec:construct-luperson-nl}

We utilize the off-the-shelf tracking algorithm \cite{zhang2020fairmot} \footnote{FairMOT: \url{https://github.com/ifzhang/FairMOT}} to detect persons and extract person tracklets from the same raw videos of \cite{fu2020unsupervised}. We assign each tracklet with a unique class label.
The detection is not perfect: \eg the bounding boxes may only cover partial bodies without heads or upper parts. 
Human pose estimation \cite{SunXLW19} is thus appended that helps filter out imperfect boxes by predicting landmarks.

We track every person in the video frame by frame. In order to guarantee both the sufficiency and diversity, we adopt the following strategy: 
i) We first remove the person identities that appear in too few frames, \ie no more than $200$;
ii) Within the tracklet of each identity, we then perform sampling with a rate of one image per $20$ frames to reduce the number of duplicated images.
Thus we can make sure that there would be at least $10$ images associating to each identity.
Through this filtering procedure, we have collected $10,683,716$ images of $433,997$ identities in total. 
They belong to $21,697$ videos which are less than the videos that \cite{fu2020unsupervised} uses, due to our extra filtering strategy for more reliable identity labels. Thus, LUPerson-NL is very different from LUPerson, as it adopts very different sampling and post-processing strategies, not to mention the noisy labels driven from the spatial-temporal information.

\begin{figure}[t]
\begin{center}
    \includegraphics[width=0.95\linewidth,height=0.34\linewidth]{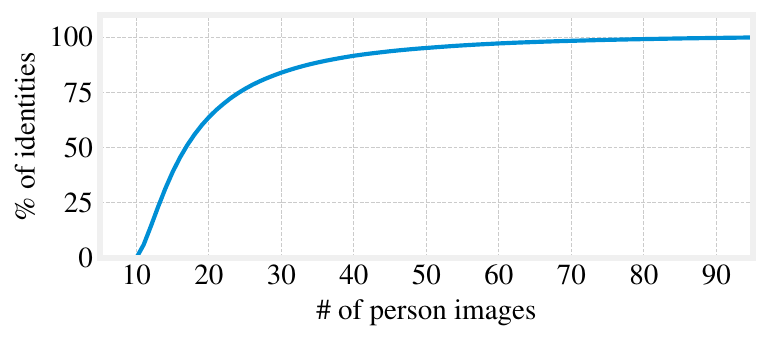}
\end{center}
\caption{Identity distribution of LUPerson-NL. A curve point $(X, Y)$ indicates $Y\%$ of identities each has less than $X$ images.}
\label{fig:hist}
\end{figure}


\begin{figure}[t]
\begin{center}
    \includegraphics[width=1.\linewidth,height=0.48\linewidth]{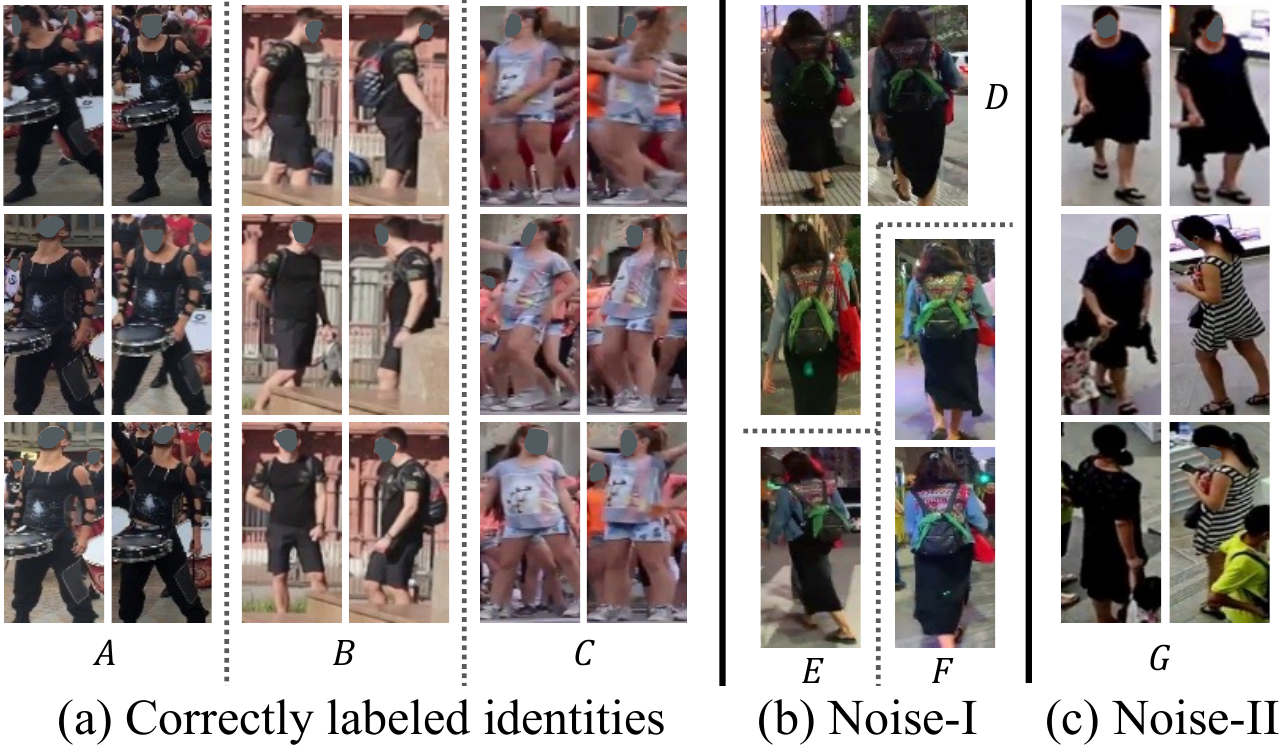}
\end{center}
\vspace{-0.7cm}
\caption{Besides the correctly labeled identities as shown by (a), there are two types of labeling errors in LUPerson-NL. \texttt{Noise-I}: same person labeled as different identities, \eg $D$, $E$ and $F$ shown in (b). \texttt{Noise-II}: different persons labeled as the same identity, \eg $G$ shown in (c).}
\label{fig:lup-ws}
\end{figure}

\subsection{Properties of LUPerson-NL}

LUPerson-NL is advantageous in following aspects:
\noindent\textbf{Large amount of images and identities.} 
We detail the statistics of existing popular person Re-ID datasets in Table \ref{tab:data-stat}.
As we can see, the proposed LUPerson-NL, with over $10M$ images and $433K$ noisy labeled identities, is the second largest among the listed. Indeed, SYSU30K has more images, but it extracts images from only \textit{1K TV program videos frame by frame}, making it less competitive in variability and less compatible in practice, the pre-training performance comparison can be found at supplementary materials. Besides, LUPerson-NL was constructed without human labeling effort, making it more suitable to scale-up. 
\noindent\textbf{Balanced distribution of identities.} 
We illustrate the cumulative percentage of identities with respect to the number of their corresponding person images as a curve in Figure \ref{fig:hist}.
A point $(X, Y)$ on the curve represents that there are in total $Y\%$ identities in LUPerson-NL, that each of them has less than $X$ images.
It can be observed that: 
i) about $75\%$ of all the identities in LUPerson-NL have a person image number within $[10, 25]$;
ii) the percentage of identities that have more than $50$ person images each, occupy only a very small portion of about $6.4\%$ ($27,767/433,997$) in LUPerson-NL.
These observations all show that our LUPerson-NL is well balanced in terms of identity distribution, making it a suitable dataset for person Re-ID tasks.

In spite of our dedicatedly designed tracking and filtering strategies as proposed in Sec \ref{ssec:construct-luperson-nl}, 
the identity labels we obtained can never be very accurate due to the technical upper bounds of current tracking methods.
Figure \ref{fig:lup-ws} visualizes the two noise types in LUPerson-NL that are caused by different labeling errors, which are
\texttt{Noise-I}, where the same person is split into different tracklets and is mistaken as different persons;
and \texttt{Noise-II}, that different persons are recognized as the same person.

\begin{figure}[t]
\begin{center}
    \includegraphics[width=1.\linewidth,height=0.57\linewidth]{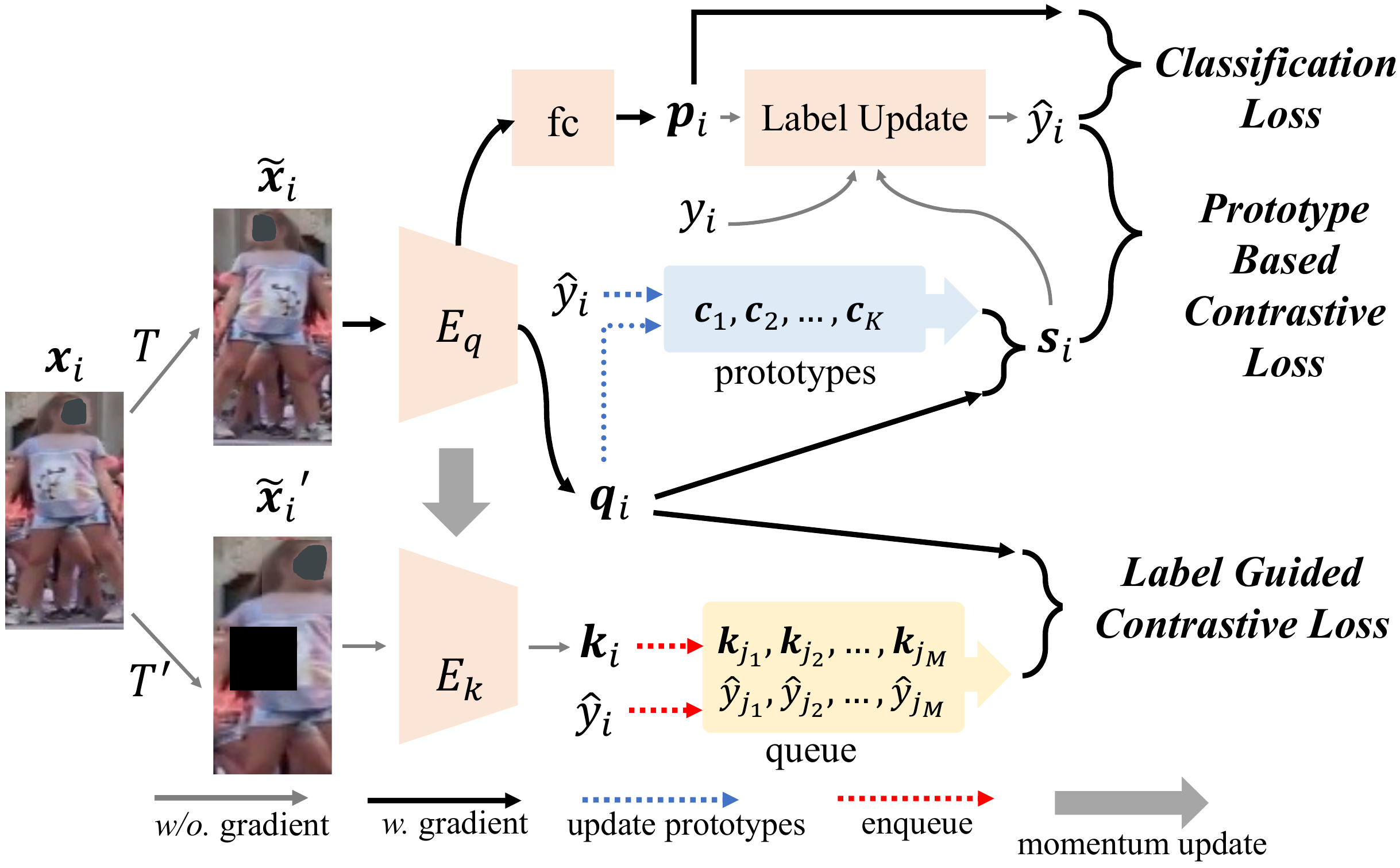}
\end{center}
\caption{The overview of our PNL framework. It comprises a supervised classification module, a prototype based contrastive learning module, and a label-guided contrastive learning module.}
\label{fig:wsp}
\end{figure}

\section{PNL: Pre-training with Noisy Labels for Person Re-ID}
Based on the new LUPerson-NL dataset with large scale noisy labels, we devise a novel \textbf{P}retraining  framework with \textbf{N}oisy \textbf{L}abels for person Re-ID, namely \textbf{PNL}. 

Denote all the data samples from LUPerson-NL as $\{(\bm{x}_i, y_i)\}^n_{i=1}$, 
with $n$ being the size of the dataset, $\bm{x}_i$ a person image and $y_i \in \{1, \dots, K\}$ its associated identity label.
Here $K$ represents the number of all identities that are recorded in LUPerson-NL.

Inspired by recent methods \cite{he2020momentum,chen2020mocov2,chen2020simple,chen2020big,grill2020bootstrap,li2020mopro}, our PNL framework adopts Siamese networks that have been fully investigated for contrastive representation learning. As shown by Figure \ref{fig:wsp},
given an input person image $\bm{x}_i$, we first perform two randomly selected augmentations $(T, {T}')$, producing two augmented images $({\tilde{\bm{x}}}_i, {\tilde{\bm{x}}}'_i)$.
We feed one of them, ${\tilde{\bm{x}}}_i$, into an encoder $E_q$ to get a \emph{query feature} $\bm{q}_i$; while the other one, ${\tilde{\bm{x}}}'_i$, is fed into another encoder $E_k$ to get a \emph{key feature} $\bm{k}_i$. Following \cite{he2020momentum}, we design $E_k$ to be a momentum version of $E_q$, \ie the two encoders $E_k$ and $E_q$ share the same network structure, but with different weights. The weights in $E_k$ are {exponential moving averages} of the weights in $E_q$. During training, weights of $E_k$ are refreshed through a momentum update from $E_q$. And the detailed algorithm can be found at supplementary materials.

\subsection{Supervised Classification}

Since the raw labels $\{y_i\}^n_{i=1}$ in LUPerson-NL contain lots of noises as illustrated in previous section, they have to be rectified during training.
Let $\hat{y}_i$ be the rectified label of image $\bm{x}_i$. As long as $\hat{y}_i$ is given, it would be intuitive that we train classification based on the corrected label $\hat{y}_i$. 
In particular, we would append a classifier to transform the feature from $E_q$ into probabilities $\bm{p}_i \in \mathbb{R}^K$ with $K$ being the number of classes. Then we impose a classification loss

\begin{equation}
\label{equ:ce}
    \mathcal{L}^i_{ce} = - \log(\bm{p}_i[\hat{y}_i]).
\end{equation}

However, the acquisition of $\hat{y}_i$ is not straight-forward. 
We resort to \emph{prototypes}, the moving averaged centroids of features from training instances, to accomplish this task.

\subsection{Label Rectification with Prototypes}
\label{ssec:label_rectification}

As depicted by Figure \ref{fig:wsp}, we maintain prototypes as a dictionary of feature vectors $\{ \bm{c}_1, \bm{c}_2, \dots, \bm{c}_K \}$, where $K$ is the number of identities, $\bm{c}_k \in \mathbb{R}^d$ is a prototype representing a class-wise feature centroid.
In each training step, we would first evaluate the similarity score $s_i^k$ between the query feature $\bm{q}_i$ and each of the current prototypes $\bm{c}_k$ by 

\begin{equation}
s_i^k = \frac{ \exp(\bm{q}_i \cdot \bm{c}_k / \tau) } {\sum_{k=1}^K {\exp(\bm{q}_i \cdot \bm{c}_k / \tau)} }.
\end{equation}

Let $\bm{p}_i$ be the classification probability given by the classifier with weights updated in the previous step.
The rectified label $\hat{y}_i$ for this step is then generated by combining both the prototype scores $\bm{s}_i = \{s_i^k\}_{k=1}^K$ and the classification probability $\bm{p}_i$ as

\begin{equation}
\label{equ:lc}
\begin{split}
    \bm{l}_i &= \frac{1}{2}(\bm{p}_i + \bm{s}_i), \\
    \hat{y}_i &= 
        \begin{cases}
        \argmax_j {\bm{l}^j_i} & \text{if~ $\max_j {\bm{l}^j_i} > T $,}\\
        y_i & \text{otherwise}.
        \end{cases}
\end{split}
\end{equation}

Here we compute a soft pseudo label $\bm{l}_i$ and convert it to a hard one $\hat{y}_i$ based on a threshold $T$. 
If the highest score in $\bm{l}_i$ is larger than $T$, the corresponding class would be selected as $\hat{y}_i$, otherwise the original raw label $y_i$ would be kept.

\subsection{Prototype Based Contrastive Learning}
\label{ssec:prototypes}

The newly rectified label $\hat{y}_i$ can then be used to supervise the cross-entropy loss $\mathcal{L}^i_{ce}$ for classification as formulated by Equation \ref{equ:ce}.
Besides, it also helps train prototypes $\bm{c}_k$ in return.
In specific, we propose a \emph{prototype based contrastive loss} $\mathcal{L}^i_{pro}$ to constrain that the feature of each sample should be closer to the prototype it belongs to.
We formulate the loss as

\begin{equation}
\label{eqn:pro}
    \mathcal{L}^i_{pro} = - \text{log}\frac{\text{exp}(\bm{q}_i \cdot \bm{c}_{\hat{y}_i} / \tau)}{\sum_{j=1}^{K} \text{exp}(\bm{q}_i\cdot\bm{c}_j/\tau)},
\end{equation}

with $\bm{q}_i$ being the query feature from $E_q$, $\tau$ being a hyper-parameter representing temperature. 

All the prototypes are maintained as a dictionary, with step-wise updates following a momentum mechanism as 

\begin{equation}
    \label{equ:pro_update}
    \bm{c}_{\hat{y}_i} = m\bm{c}_{\hat{y}_i} + (1 - m)\bm{q}_i.
\end{equation}

\subsection{Label-Guided Contrastive Learning}
\label{ssec:ccl}

Instance-wise contrastive learning proved to be very effective in self-supervised learning \cite{he2020momentum,chen2020mocov2,chen2020simple,chen2020big,grill2020bootstrap}.
It learns instance-level feature discrimination by encouraging similarity among features from the same instance, while promoting dissimilarity between features from different instances. 
The instance-wise contrastive loss is given by

\begin{equation}
\label{eqn:cont-inst}
    \mathcal{L}^i_{ic} = - \text{log}\frac{\text{exp}(\bm{q}_i\cdot\bm{k}^+_i / \tau)}{\text{exp}(\bm{q}_i\cdot\bm{k}^+_i/\tau) + \sum_{j=1}^{M} \text{exp}(\bm{q}_i\cdot\bm{k}^-_j/\tau)},
\end{equation}

with $\bm{q}_i$ being the query feature of current instance $i$. 
$\bm{k}_i^+ (= \bm{k}_i)$ is the positive key feature generated from the momentum encoder $E_k$. It is marked \emph{positive} since it shares the same instance with $\bm{q}_i$.
$\bm{k}_*^- \in \mathbb{R}^d$, on the contrary, are the rest features stored in a queue that represent \emph{negative} samples. 
The queue has a size of $M$. At the end of each training step, the queue would be updated by en-queuing the new key feature and de-queuing the oldest one.

Such instance-level contrastive learning is far from perfect, as it neglects the relationships among different instances. 
For example, even though two instances depict the same person, it would still strengthen the gap between their features.
Instead, we propose a \emph{label guided contrastive learning} module, making use of the rectified labels $\hat{y}_i$ to ensure more reasonable grouping of contrastive pairs.

We redesign the queue to additionally record labels $\hat{y}_i$. 
Represented by $\mathcal{Q} = [(\bm{k}_{j_t}, \hat{y}_{j_t})]_{t=1}^M $, our new queue accepts not only a key feature $\bm{k}_i$ but also its rectified label $\hat{y}_i$ during update.
These newly recorded labels help better distinguish positive and negative pairs.
Let $\mathcal{P}(i)$ be the new set of positive features and $\mathcal{N}(i)$ the new set of negative features: features in $\mathcal{P}(i)$ share the same rectified label with the current instance $i$ while features in $\mathcal{N}(i)$ do not.
Our label guided contrastive loss can be given by

\begin{equation}
\label{equ:label-guided-contrast}
\mathcal{L}^i_{lgc} = \frac{-1}{|\mathcal{P}(i)|} \log
\frac{ \sum\limits_{\mathclap{\bm{k}^+ \in \mathcal{P}(i)}}{} \exp\left( \frac{\bm{q}_i \cdot \bm{k}^+} {\tau} \right) }
	 { \sum\limits_{\mathclap{\bm{k}^+ \in \mathcal{P}(i)}}{} \exp\left( \frac{\bm{q}_i \cdot \bm{k}^+} {\tau} \right) 
	 + \sum\limits_{\mathclap{\bm{k}^- \in \mathcal{N}(i)}}{} \exp\left( \frac{\bm{q}_i \cdot \bm{k}^-} {\tau} \right)},
\end{equation}
with 
\begin{align}
\mathcal{P}(i) &= \{ \bm{k}_{j_t} | \hat{y}_{j_t}=\hat{y}_i, \forall (\bm{k}_{j_t}, \hat{y}_{j_t}) \in \mathcal{Q} \} \cup \{ \bm{k}_i \},\nonumber\\
\mathcal{N}(i) &= \{ \bm{k}_{j_t} | \hat{y}_{j_t}\neq\hat{y}_i, \forall (\bm{k}_{j_t}, \hat{y}_{j_t}) \in \mathcal{Q}  \},
\end{align}
where $\bm{k}_i$ and $\hat{y}_i$ are the key feature and the rectified label of the current instance $i$. 

Finally we combine all the components above to pre-train models on LUPerson-NL with the following loss 

\begin{equation}
    \mathcal{L}^i = \mathcal{L}^i_{ce} + \lambda_{pro}\mathcal{L}^i_{pro} + \lambda_{lgc}\mathcal{L}^i_{lgc}.
\end{equation}
We set $\lambda_{pro}= \lambda_{lgc} = 1$ during training.

\begin{table*}[htb]
\setlength{\tabcolsep}{3.3mm}
    \begin{subtable}[h]{0.5\textwidth}
        \centering
        \begin{tabular}{l|ccc}
        \shline
        pre-train & Trip~\cite{hermans2017defense} & IDE~\cite{zheng2017person} & MGN~\cite{wang2018learning} \\
        \hline
        IN sup.     & 45.2/63.8 & 50.6/55.9 & 70.5/71.2 \\
        IN unsup.   & 55.5/61.2 & 52.5/57.7 & 67.1/67.0 \\ 
        LUP unsup.  & 62.6/67.6 & 57.6/62.3 & 74.7/75.4 \\ 
        \hline
        LUPnl pnl. & \textbf{69.1/73.1} & \textbf{68.3/73.5} & \textbf{80.4/80.9} \\
        \shline
        \end{tabular}
        \caption{CUHK03}
        \label{tab:improve-cuhk}
    \end{subtable}
    \hfill
    \begin{subtable}[h]{0.5\textwidth}
        \centering
        \begin{tabular}{l|ccc}
        \shline
        pre-train & Trip~\cite{hermans2017defense} & IDE~\cite{zheng2017person} & MGN~\cite{wang2018learning} \\
        \hline
        IN sup.    & 76.2/89.7 & 74.1/90.2 & 87.5/95.1 \\
        IN unsup.  & 75.1/88.5 & 74.5/89.3 & 88.2/95.3 \\ 
        LUP unsup. & 79.8/71.5 & 77.9/91.0 & 91.0/96.4 \\ 
        \hline
        LUPnl pnl. & \textbf{81.2/91.4} & \textbf{82.4/92.8} & \textbf{91.9/96.6} \\
        \shline
        \end{tabular}
        \caption{Market1501}
        \label{tab:improve-market}
    \end{subtable}
    \hfill
    \begin{subtable}[h]{0.5\textwidth}
        \centering
        \begin{tabular}{l|ccc}
        \shline
        pre-train & Trip~\cite{hermans2017defense} & IDE~\cite{zheng2017person} & MGN~\cite{wang2018learning} \\
        \hline
        IN sup.    & 65.2/80.7 & 62.8/80.8 & 79.4/89.0 \\
        IN unsup.  & 65.4/81.1 & 63.4/81.6 & 79.5/89.1 \\ 
        LUP unsup. & 69.8/83.1 & 65.9/82.2 & 82.1/91.0 \\ 
        \hline
        LUPnl pnl. & \textbf{71.0/84.7} & \textbf{70.3/85.0} & \textbf{84.3/92.0} \\
        \shline
        \end{tabular}
        \caption{DukeMTMC}
        \label{tab:improve-duke}
    \end{subtable}
    \hfill
    \begin{subtable}[h]{0.5\textwidth}
        \centering
        \begin{tabular}{l|ccc}
        \shline
        pre-train & Trip~\cite{hermans2017defense} & IDE~\cite{zheng2017person} & MGN~\cite{wang2018learning} \\
        \hline
        IN sup.    & 34.3/54.8 & 36.2/66.2 & 63.7/85.1 \\
        IN unsup.  & 34.4/55.4 & 37.6/67.3 & 62.7/84.3 \\ 
        LUP unsup. & 36.6/57.1 & 39.8/68.9 & 65.7/85.5 \\
        \hline
        LUPnl pnl. & \textbf{41.4/61.6} & \textbf{44.0/72.0} & \textbf{68.0/86.0} \\
        \shline
        \end{tabular}
        \caption{MSMT17}
        \label{tab:improve-msmt}
    \end{subtable}
    \vspace{-0.3cm}
    \caption{Comparing three supervised Re-ID baselines using different pre-trained models. 
    ``IN sup."/``IN unsup." indicates model that is supervisely/unsupervisely pre-trained on ImageNet; ``LUP unsup." is the model unsupervisely pre-trained on LUPerson; 
    ``LUPnl pnl.`` refers to the model that pre-trained on LUPerson-NL using our PNL framework. 
    All results are shown in \emph{mAP}/\emph{cmc1}.}
    \label{tab:impro-sup}
    \vspace{-0.8em}
\end{table*}

\section{Experiments}

\begin{table*}[h]
\setlength{\tabcolsep}{1.7mm}
\small
    \begin{subtable}[h]{1.0\textwidth}
        \centering
        \begin{tabular}{l|ccccc|ccccc}
        \shline
        \multirow{2}{*}{pre-train} & \multicolumn{5}{c|}{small-scale} & \multicolumn{5}{c}{few-shot} \\ \cline{2-11} & $10\%$ & $30\%$ & $50\%$ & $70\%$ & $90\%$ & $10\%$ & $30\%$ & $50\%$ & $70\%$ & $90\%$ \\ \hline
        IN sup.    & 53.1/76.9 & 75.2/90.8 & 81.5/93.5 & 84.8/94.5 & 86.9/95.2 & 21.1/41.8 & 68.1/87.6 & 80.2/92.8 & 84.2/94.0 & 86.7/94.6 \\
        IN unsup.  & 58.4/81.7 & 76.6/91.9 & 82.0/94.1 & 85.4/94.5 & 87.4/95.5 & 18.6/36.1 & 69.3/87.8 & 78.3/90.9 & 84.4/94.1 & 87.1/95.2 \\
        LUP unsup. & 64.6/85.5 & 81.9/93.7 & 85.8/94.9 & 88.8/95.9 & 90.5/96.4 & 26.4/47.5 & 78.3/92.1 & 84.2/93.9 & 88.4/95.5 & 90.4/96.3 \\
        \hline
        LUPnl pnl. & \textbf{72.4/88.8} & \textbf{85.2/94.2} & \textbf{88.3/95.5} & \textbf{90.1/96.2} & \textbf{91.3/96.4} & \textbf{42.0/61.6} & \textbf{83.7/94.0} & \textbf{88.1/95.2} & \textbf{90.5/96.3} & \textbf{91.6/96.4} \\
        \shline
    \end{tabular}
    \caption{Market1501}
    \label{tab:sd-fw-market}
    \end{subtable}
    
   \begin{subtable}[h]{1.0\textwidth}
        \centering
        \begin{tabular}{l|ccccc|ccccc}
        \shline
        \multirow{2}{*}{pre-train} & \multicolumn{5}{c|}{small-scale} & \multicolumn{5}{c}{few-shot} \\ \cline{2-11} & $10\%$ & $30\%$ & $50\%$ & $70\%$ & $90\%$ & $10\%$ & $30\%$ & $50\%$ & $70\%$ & $90\%$ \\ \hline
        IN sup.    & 45.1/65.3 & 64.7/80.2 & 71.8/84.6 & 75.5/86.8 & 78.0/88.3 & 31.5/47.1 & 65.4/79.8 & 73.9/85.7 & 77.2/87.8 & 79.1/88.8 \\ 
        IN unsup.  & 48.1/66.9 & 65.8/80.2 & 72.5/84.4 & 76.3/86.9 & 78.5/88.7 & 32.4/48.0 & 65.3/80.2 & 73.7/85.1 & 77.7/87.8 & 79.4/89.0 \\
        LUP unsup. & 53.5/72.0 & 69.4/81.9 & 75.6/86.7 & 78.9/88.2 & 81.1/90.0 & 35.8/50.2 & 72.3/83.8 & 77.7/87.4 & 80.8/89.2 & 82.0/90.6 \\
        \hline
        LUPnl pnl. & \textbf{60.6/75.8} & \textbf{74.5/86.3} & \textbf{78.8/88.3} & \textbf{81.6/89.5} & \textbf{83.3/91.2} & \textbf{52.2/64.1} & \textbf{77.7/87.9} & \textbf{81.1/89.6} & \textbf{83.2/91.1} & \textbf{84.1/91.3} \\
        \shline
    \end{tabular}
    \caption{DukeMTMC}
    \label{tab:sd-fw-duke}
    \end{subtable}
    
    \begin{subtable}[h]{1.0\textwidth}
        \centering
        \begin{tabular}{l|ccccc|ccccc}
        \shline
        \multirow{2}{*}{pre-train} & \multicolumn{5}{c|}{small-scale} & \multicolumn{5}{c}{few-shot} \\ \cline{2-11} & $10\%$ & $30\%$ & $50\%$ & $70\%$ & $90\%$ & $10\%$ & $30\%$ & $50\%$ & $70\%$ & $90\%$ \\ \hline
        IN sup.    & 23.2/50.2 & 41.9/70.8 & 50.3/76.9 & 56.9/81.2 & 61.9/84.2 & 14.7/34.1 & 44.5/71.1 & 56.2/79.5 & 60.9/82.8 & 63.4/84.5 \\ 
        IN unsup.  & 22.6/48.8 & 40.4/68.7 & 49.0/75.0 & 55.7/79.9 & 60.9/83.0 & 13.2/29.2 & 41.4/67.1 & 53.3/77.6 & 59.1/81.5 & 62.4/83.8 \\
        LUP unsup. & 25.5/51.1 & 44.6/\textbf{71.4} & 53.0/\textbf{77.7} & 59.5/81.8 & 63.7/\textbf{85.0} & 17.0/36.0 & 49.0/73.6 & 57.4/80.5 & 62.9/83.5 & 65.0/85.1 \\
        \hline
        LUPnl pnl. & \textbf{28.2/51.1} & \textbf{47.7}/71.2 & \textbf{55.5}/77.2 & \textbf{61.6/81.8} & \textbf{66.1}/84.8 & \textbf{24.5/42.7} & \textbf{53.2/74.4} & \textbf{62.2/81.0} & \textbf{65.8/83.8} & \textbf{67.4/85.3} \\
        \shline
    \end{tabular}
    \caption{MSMT17}
    \label{tab:sd-fw-msmt}
    \vspace{-1em}
    \end{subtable}
    \caption{Comparing pre-trained models on three labeled Re-ID datasets, under the \emph{small-scale} setting and the \emph{few-shot} setting, with different usable data percentages. ``LUPnl pnl.'' is our model pre-trained on LUPerson-NL using PNL. Results are shown in  \emph{mAP}/\emph{cmc1}.}
    \label{tab:sd-fw}
    \vspace{-0.9em}
\end{table*}

\subsection{Implementation}
\noindent\textbf{Hyper-parameter settings.} We set the hyper-parameters $\tau=0.1$ and $T=0.8$. The momentum $m$ for updating both the momentum encoder $E_k$ and the prototypes is set to $0.999$. More hyper-parameters exploration and training details can be found at supplementary materials. 

\noindent\textbf{Dataset and protocol.} 
We conduct extensive experiments on four popular person Re-ID datasets: CUHK03, Market, DukeMTMC and MSMT17. 
We adopt their official settings, except CUHK03 where its labeled counterpart with new protocols proposed in~\cite{zhong2017re} is used. 
We follow the standard evaluation metrics: the mean Average Precision (mAP) and the Cumulated Matching Characteristics top-1 (cmc1).

\subsection{Improving Supervised Re-ID}

To evaluate our pre-trained model based on LUPerson-NL with respect to supervised person Re-ID tasks.
we conduct experiments using three representative supervised Re-ID baselines with different pre-training models. 
These baseline methods include two simpler approaches driven only by the triplet loss (Trip~\cite{hermans2017defense}) or the classification loss (IDE~\cite{zheng2017person}), as well as a stronger and more complex method MGN~\cite{wang2018learning} that use both triplet and classification losses.

We report results in Table \ref{tab:impro-sup}, where the abbreviations \{``IN'', ``LUP'', ``LUPnl''\} represent ImageNet \cite{russakovsky2015imagenet}, LUPerson \cite{fu2020unsupervised} and our LUPerson-NL respectively; while the \{``sup.'', ``unsup.'', ``pnl.''\} stand for the \{``supervised'', ``unsupervised'', and ``pretrain with noisy label''\} pre-training methods. \eg the \textbf{``LUPnl pnl.''} in the bottom rows of Table \ref{tab:impro-sup} all refer to our model, which is pre-trained on our LUPerson-NL dataset using our PNL framework.

From Table \ref{tab:impro-sup} we can see, for all of the three baseline methods, our pre-trained model improves their performances greatly on the four popular person Re-ID datasets. Specifically, the improvements are at least 5.7\%, 0.9\%, 1.2\% and 2.3\% in terms of \emph{mAP} on CUHK03, Market1501, DukeMTMC and MSMT17 respectively. 

Note that even though the performance of the baseline MGN on Market1501 has been extremely high, our model still brings considerable improvement over it.
The other way around, our pre-trained models obtain more significant improvements on relatively weak methods (Trip and IDE),
unveiling that model initialization plays a critical part in person Re-ID training.

Our noisy label guided pre-training models are also significantly advantageous over the previous {``LUPerson unsup``} models, which emphasizes the superiority of our PNL framework and our LUPerson-NL dataset.  

\subsection{Improving Unsupervised Re-ID Methods}

Our pre-trained model can also benefit unsupervised person Re-ID methods. 
Based on the state-of-the-art unsupervised method SpCL~\cite{ge2020selfpaced}, 
we explore different pre-training models utilizing two settings proposed by SpCL: the pure unsupervised learning (USL) and the unsupervised domain adaptation (UDA).
Results in Table \ref{tab:impor-unsup} illustrate that our pre-trained model outperforms the others in all UDA tasks, as well as the USL task on DukeMTMC dataset. 
In the USL task on Market1501, we achieve the second best scores slightly lower than the LUPerson model \cite{fu2020unsupervised}.

\subsection{Comparison on Small-scale and Few-shot}

Following the same protocols proposed by \cite{fu2020unsupervised}, we conduct experiments under two small data settings: the \emph{small-scale} setting and the \emph{few-shot} setting.
The small-scale setting restricts the percentage of usable identities, while the few-shot setting restricts the percentage of usable person images each identity has.
Under both settings, we vary the usable data percentages of three popular datasets from $10\%\sim100\%$. 
We compare different pre-trained models under these settings with MGN as the baseline method.
The results shown in Table \ref{tab:sd-fw} verify the consistent improvements brought by our model on all the datasets under both settings.

Besides, the results in Table~\ref{tab:sd-fw} show that the gains of our pre-trained models are even larger under a {more limited amount of labeled data}.
For example, under the ``small-scale'' setting, our model outperforms ``LUPerson unsup'' by 7.8\%, 7.1\% and 2.7\% on Market1501, DukeMTMC and MSMT17 respectively with $10\%$ identities.
The improvements rise to 15.6\%, 16.4\% and 6.5\% under the ``few-shot'' setting with $10\%$ person images. 

Most importantly, our pre-trained ``LUPnl pnl'' model helps achieve advantageous results with a \emph{mAP} of $72.4$ and a \emph{cmc1} of $88.8$, using only $10\%$ labeled data from the Market1501 training set. The task is really challenging, considering that the training set composes only $1,170$ images belonging to $75$ identities; while evaluations are performed on a much larger testing set with $19,281$ images belonging to $750$ identities.
We consider these results extremely appealing as they demonstrate the strong potential of our pre-trained models in real-world applications.

\begin{table}[t]
\small
    \centering
    \begin{tabular}{l|cc|cc}
    \shline
    \multirow{2}{*}{pre-train} & \multicolumn{2}{c|}{USL} & \multicolumn{2}{c}{UDA} \\ 
    \cline{2-5} & M & D & D $\rightarrow$ M & M $\rightarrow$ D \\ 
    \hline
    IN sup.    & 72.4/87.8 & 64.9/80.3 & 76.4/90.1 & 67.9/82.3 \\
    IN unsup.  & 72.9/88.6 & 62.6/78.8 & 77.1/90.6 & 66.3/81.6 \\
    LUP unsup. & \textbf{76.2/90.2} & 67.1/81.6 & 79.2/91.7 & 69.1/83.2 \\
    \hline
    LUPnl pnl. & 75.6/89.3 & \textbf{68.1/82.0} & \textbf{80.7/92.2} & \textbf{72.2/84.9} \\
    \shline
\end{tabular}
\caption{Performances of different pre-trained models on the unsupervised Re-ID method SpCL \cite{ge2020selfpaced} under two unsupervised task settings: the pure unsupervised learning (USL) and the unsupervised domain adaptation (UDA). Here M and D refer to the Market1501 dataset and the DukeMTMC dataset respectively.}
\label{tab:impor-unsup}
\end{table}

\subsection{Comparison with other pre-training methods}
We compare our proposed PNL with some other popular pre-training methods in Table~\ref{tab:com_method}. LUP~\cite{fu2020unsupervised} is a varient of MoCoV2 for person Re-ID based on unsupervised constrastive learning, while SupCont~\cite{khosla2020supervised} considers both supervised and constrastive learning. Our PNL outperforms all these rep-resentative pre-training methods, indicating the superiority of our proposed method.
\begin{table}[t]
\setlength{\tabcolsep}{2.6mm}
    \centering
    \vspace{-0.2cm}
    \begin{tabular}{l|c|c|c}
        \shline
       method  &  SupCont~\cite{khosla2020supervised} & LUP~\cite{fu2020unsupervised} & PNL(ours) \\
       \hline
       MSMT17  &  66.5/84.7 & 65.3/84.0 & 68.0/86.0 \\
       \shline
    \end{tabular}
    \caption{Performance comparison for different pre-training methods on LUPerson-NL dataset.}
    \label{tab:com_method}
\end{table}

\begin{table}[t]
\setlength{\tabcolsep}{1.6mm}
\centering
    \begin{tabular}{l|llll|c|c|c}
    \shline
    \# & ${ce}$ & ${ic}$ & ${pro}$ & ${lgc}$ & 20\% & 40\% & 100\% \\ 
    \hline
    1 & \checkmark &  &  &  & 32.0/56.1 & 45.0/69.5 & 62.7/83.0 \\
    2 & & \checkmark &   &  & 34.5/59.5 & 47.9/72.6 & 65.3/84.0 \\
    3 & \checkmark & \checkmark &  &  & 37.6/62.6 & 49.6/73.5 & 66.5/84.7 \\
    4 & \checkmark & & \checkmark & & 35.7/59.1 & 48.5/72.4 & 65.8/84.1 \\
    5 & \checkmark & & & \checkmark & 38.5/63.0 & 50.9/74.5 & 67.1/85.2 \\
    6 & \checkmark & \checkmark & \checkmark &  & 39.0/63.4 & 51.7/74.4 & 67.4/85.4 \\
    7 & \checkmark &  & \checkmark & \checkmark & \textbf{39.6/63.7} & \textbf{51.9/75.0} & \textbf{68.0/86.0} \\
    \shline
\end{tabular}
\caption{Ablating components of PNL on MSMT with data percentages $20\%$, $40\%$ and $100\%$ under the small scale setting. $ce$: supervised classification; $ic$: instance-wise contrastive learning; $pro$: prototypes for both prototype-based contrastive learning and label rectification; $lgc$: label-guided contrastive learning.}
\label{tab:ablation}
\end{table}

\subsection{Ablation Study}

We also investigate the effectiveness of each designed component in PNL through ablation experiments. Results shown by Table \ref{tab:ablation} illustrate the efficacy of our proposed components. We have the following observations:
\textbf{i)} Training with an instance-wise contrastive loss $\mathcal{L}_{ic}^i$ (row 2) without using any labels leads to even better performance than training with a classification loss $\mathcal{L}_{ce}^i$ (row 1) that utilizes the labels from LUPerson-NL, implying that the noisy labels in LUPerson-NL would misguide representation learning if directly adopted as supervision.
\textbf{ii)} Jointly training with both losses $\mathcal{L}_{ce}^i$ and $\mathcal{L}_{ic}^i$ (row 3) improves over using only one loss (row 1, row 2), suggesting that learning instance-wise discriminative representations complements label supervision.
\textbf{iii)} The prototypes which contribute to both prototype-based contrastive learning and the label correction, are very important under various settings, as verified by comparing row 1 with row 4; row 3 with row 6; and row 5 with row 7.
\textbf{iv)} Our label-guided contrastive learning component consistently outperforms the vanilla instance-wise contrastive learning under various settings, as verified by comparing row 3 with row 5, and row 6 with row 7.
\textbf{v)} Combining all our components (supervised classification, prototypes and label-guided contrastive learning) together leads to the best performance as shown by row 7.

\begin{figure}[t]
\begin{center}
    \begin{subfigure}{3.3in}
        \includegraphics[width=3.3in]{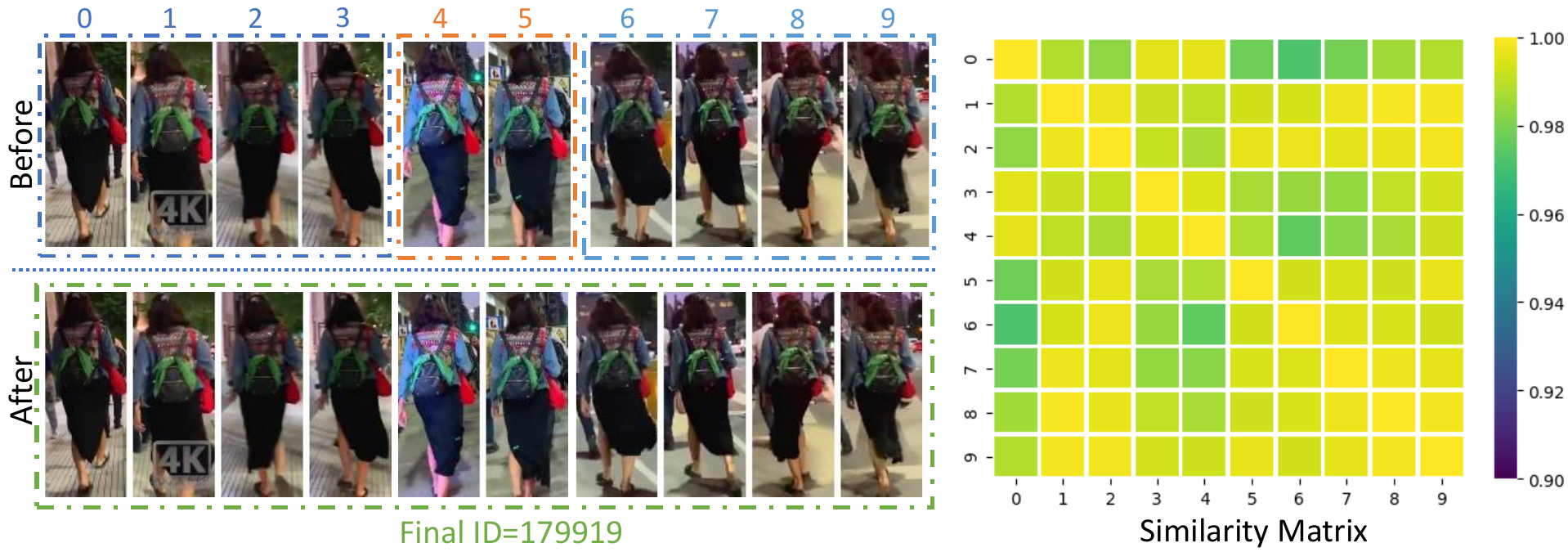}
        \caption{Correction for \texttt{Noise-I}}
        \label{fig:vis-lc-i}
    \end{subfigure}
    \begin{subfigure}{3.25in}
        \includegraphics[width=3.3in]{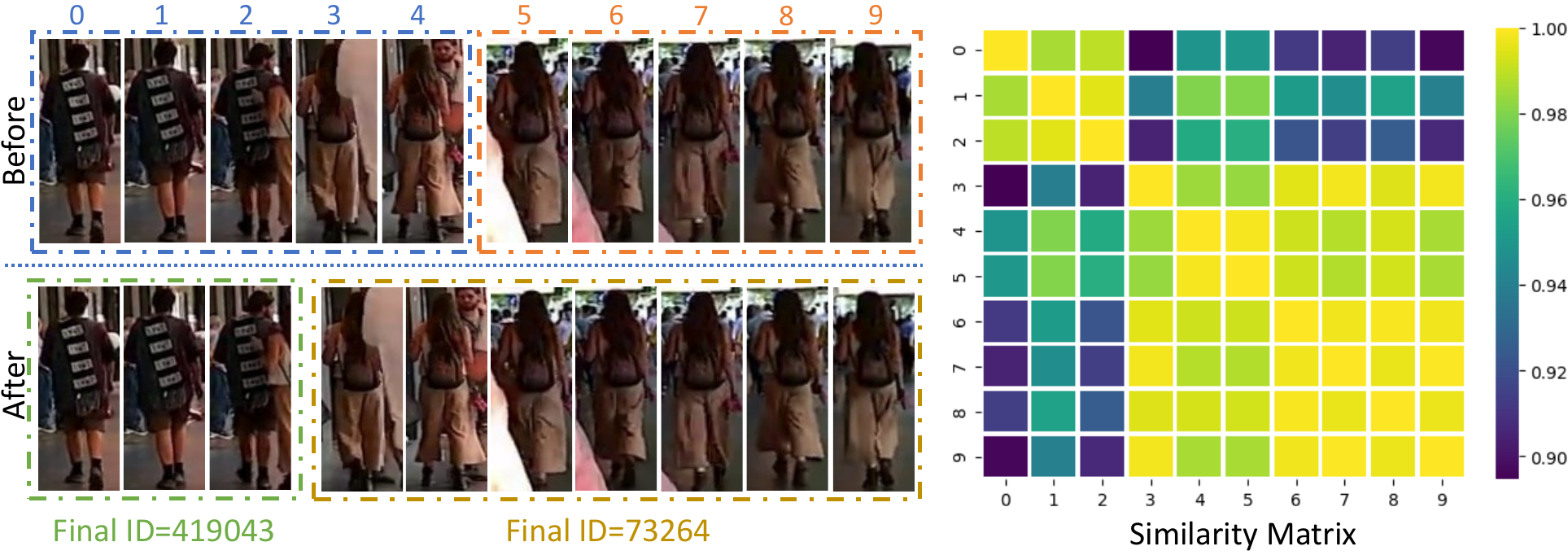}
        \caption{Correction for \texttt{Noise-II}}
        \label{fig:vis-lc-ii}
    \end{subfigure}    
\end{center}
\caption{Visualizing the label correction functionality of our PNL framework with respect to the two noise types from LUPerson-NL. Person images in the same rectangle indicate that they are  recognized as the same identity. The right-hand similarity matrices are calculated based on the image features all learned using our PNL framework with label correction.}
\label{fig:vis-lc}
\end{figure}

\subsection{Label Correction}
Our PNL can indeed correct noisy labels. We demonstrate two typical examples in Figure~\ref{fig:vis-lc} visualizing the label corrections with respect to the two kinds of noises. 
As we can see, in Figure~\ref{fig:vis-lc-i} the same person are marked as three different persons in LUPerson-NL due to \texttt{Noise-I} in labels. 
After our PNL pre-training, these three tracklets are merged together since their trained features are very close, as verified by the right-hand similarity matrix.
In Figure~\ref{fig:vis-lc-ii} different persons are labeled as the same identity in LUPerson-NL due to \texttt{Noise-II} in labels. 
After PNL training, these mis-labeled person identities are all correctly re-grouped into two identities, which can also be reflected by the right-hand similarity matrix. 

we also ablate the label correction module in Table~\ref{tab:lc} with different settings, and observe it can improve the performance. It also validates the importance of combining label rectification with label-guided contrastive learning together, where more accurate positive/negative pairs can be leveraged.

\begin{table}[t]
\setlength{\tabcolsep}{2.8mm}
    \centering
    \footnotesize
    \begin{tabular}{l|c|c|c|c}
    \shline
    \multirow{2}{*}{setting} & \multicolumn{2}{c|}{$ce$+$pro$} & \multicolumn{2}{c}{$ce$+$pro$+$lgc$} \\
    \cline{2-5} & w/o. $lc$ & w. $lc$       & w/o. $lc$         & w. $lc$         \\ 
    \hline
    MSMT17  & 64.8/83.4  & 65.8/84.1   & 66.7/85.0 & 68.0/86.0  \\
    \shline
    \end{tabular}
    \caption{Ablating the label correction. $lc$: ``label correction''.}
    \label{tab:lc}
\end{table}

\vspace{-0.5em}
\subsection{Comparison with State-of-the-Art Methods}
\vspace{-0.5em}
We compare our results with current state-of-the-art methods on four public benchmarks. 
We don't apply any post-processing techniques such as IIA~\cite{fu2020improving} and RR~\cite{zhong2017re}. 
To ensure fairness, we adopt ResNet50 as our backbone and does not compare with methods that rely on stronger backbones (results with stronger backbones \eg ResNet101 can be found at supplementary materials). Results in Table~\ref{tab:comp-sota} verify the remarkable advantage brought by our pre-trained models.
Without bells and whistles, we achieve state-of-the-art performance on all four benchmarks, outperforming the second with clear margins.

\begin{table}
\footnotesize
\setlength{\tabcolsep}{1.1mm}
    \centering
    \begin{tabular}{l|cccc}
    \shline
    Method & CUHK03 & Market1501 & DukeMTMC & MSMT17 \\ 
    \hline
    MGN$\dag$~\cite{wang2018learning} (2018) & 70.5/71.2 & 87.5/95.1 & 79.4/89.0 & 63.7/85.1 \\
    BOT~\cite{luo2019strong} (2019) & - & 85.9/94.5 & 76.4/86.4 & - \\
    DSA~\cite{zhang2019densely} (2019) & 75.2/78.9 & 87.6/95.7 & 74.3/86.2 & - \\
    Auto~\cite{quan2019auto} (2019) & 73.0/77.9 & 85.1/94.5 & - & 52.5/78.2 \\
    ABDNet~\cite{chen2019abd} (2019) & - & 88.3/95.6 & 78.6/89.0 & 60.8/82.3 \\
    SCAL~\cite{chen2019self} (2019) & 72.3/74.8 & 89.3/95.8 & 79.6/89.0 & - \\
    MHN~\cite{chen2019mixed} (2019) & 72.4/77.2 & 85.0/95.1 & 77.2/89.1 & - \\
    BDB~\cite{dai2019batch} (2019) & 76.7/79.4 & 86.7/95.3 & 76.0/89.0 & - \\
    SONA~\cite{xia2019second} (2019) & 79.2/81.8 & 88.8/95.6 & 78.3/89.4 & - \\
    GCP~\cite{park2020relation} (2020) & 75.6/77.9 & 88.9/95.2 & 78.6/87.9 & - \\
    SAN~\cite{jin2020semantics} (2020) & 76.4/80.1 & 88.0/96.1 & 75.5/87.9 & 55.7/79.2 \\
    ISP~\cite{zhu2020identity} (2020) & 74.1/76.5 & 88.6/95.3 & 80.0/89.6 & - \\
    GASM~\cite{he2020guided} (2020) & - & 84.7/95.3 & 74.4/88.3 & 52.5/79.5 \\
    ESNET~\cite{shen2020net} (2020) & - & 88.6/95.7 & 78.7/88.5 & 57.3/80.5 \\
    LUP~\cite{fu2020unsupervised}(2020) & \underline{79.6/81.9*} & \underline{91.0/96.4} & \underline{82.1/91.0} & \underline{65.7/85.5} \\
    \hline
    Ours+MGN & 80.4/80.9 & \textbf{91.9}/\textbf{96.6} & \textbf{84.3}/\textbf{92.0} & \textbf{68.0}/\textbf{86.0} \\
    Ours+BDB & \textbf{82.3/84.7} & 88.4/95.4 & 79.0/89.2 & 53.4/79.0 \\
    \shline
\end{tabular}
\caption{Comparison with the state of the art. 
Numbers of MGN$\dag$ come from a re-implementation based on FastReID, which are even better than the original. 
Numbers of PNL marked by * are obtained on BDB, the rest without the * mark are obtained on MGN. 
We show best scores in bold and the second scores underlined.}
\label{tab:comp-sota}
\end{table}

\section{Conclusion}
In this paper, we demonstrate that large-scale Re-ID representation can be directly learned from massive raw videos by leveraging the spatial and temporal information. We not only build a large-scale noisy labeled person Re-ID dataset \textbf{LUPerson-NL} based on tracklets of raw videos from LUPerson without using manual annotations, but also design a novel weakly supervised pretraining framework \textbf{PNL} comprising different learning modules including supervised learning, prototypes-based learning, label-guided contrastive learning and label rectification. Equipped with our pre-trained models, we push existing benchmark results to a new limit, which outperforms unsupervised pre-trained models and ImageNet supervised pre-trained models by a large margin.

\noindent \textbf{Acknowledgement.} This work is partially supported by the National Natural Science Foundation of China (NSFC, 61836011).

{\small
\bibliographystyle{ieee_fullname}
\bibliography{egbib}
}

\clearpage
\newpage
\appendix
In this material, we will 1) show demo cases for scene changing for specific person in LUPerson-NL, 2) share the training details for PNL, 3) provide the detailed algorithm table for PNL, 4) compare models pre-trained on SYSU30K and LUPerson-NL 5) analyze the hyper parameters of our PNL, 6) demonstrate results of our method using a stronger backbones, 7) explore the impact of pre-training image scale, and 8) list more detailed results for our ``small-scale" and ``few-shot" experiments. All these experiments are under MGN settings.

\section{Scene changing in LUPerson-NL}
\begin{figure}[ht]
	\begin{center}
		\includegraphics[width=\linewidth]{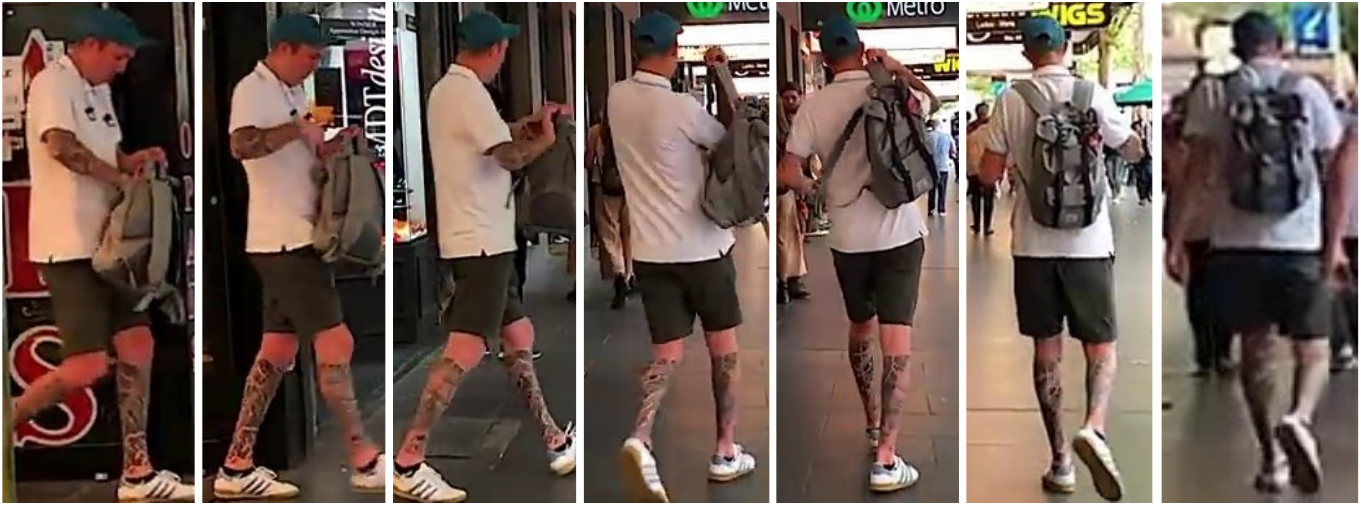}
	\end{center}
	\caption{Scene changing for a specific person in LUPerson-NL.}
	\label{fig:scene-change}
\end{figure}

Our LUPerson-NL are driven from street view videos, Figure~\ref{fig:scene-change} shows a demo person in our LUPerson-NL. As we can see, our LUPerson-NL is able to cover multiple scenes, since the videos we use have lots of moving cameras and moving persons, and only traklets with $\geq 200$ frames are selected. It is approximate close to the real person Re-ID scenario.

\section{Training details} 
During training, all images are resized to $256\times128$ and pass through the same augmentations verified in [\textcolor{green}{12}], which are Random Resized Crop, Horizontal Flip, Normalization, Random Gaussian Blur, Random Gray Scale and Random Erasing. 
Specifically, the images are normalized with mean and std of $[0.3452, 0.3070, 0.3114],[0.2633, 0.2500, 0.2480]$, which are calculated from all images in LUPerson-NL. 
We train our model on $8\times V100$ GPUs for $90$ epochs with a batch size of $1,536$.
The initial learning rate is set to $0.4$ with a step-wise decay by $0.1$ for every $40$ epochs. 
The optimizer is SGD with a momentum of $0.9$ and a weight decay of $0.0001$. 
We set the hyper-parameters $\tau=0.1$ and $T=0.8$. 
The momentum $m$ for updating both the momentum encoder $E_k$ and the prototypes is set to $0.999$. 
We design a large queue with a size of $65,536$ to increase the occurrence of positive samples for the label guided contrastive learning. 
The label correction according to Equation \textcolor{red}{3} starts from the $10$-th epoch. 
The deployment of the label guided contrastive loss $\mathcal{L}_{lgc}^i$ starts from the $15$-th epoch.

\section{Algorithm for PNL}
Algorithm~\ref{alg:pnl} shows the procedure of training PNL, we train our framework for 90 epochs, and apply label correction from 10 epochs. Once the rectification begins, we keep rectifying for every iteration.

\begin{algorithm}[ht]
	{
		\textbf{Input:} total epochs $N$, correction start epochs $N_s$, prototype feature vectors $\{ \Vec{c}_1, \Vec{c}_2, \dots, \Vec{c}_K \}$, where $K$ is the number of identities, temperature $\tau$, threshold $T$, momentum $m$, encoder network $E_q(\cdot)$, classifier $h(\cdot)$, momentum encoder $E_k(\cdot)$, loss weight $\lambda_{pro}$ and $\lambda_{lgc}$.
		
		\For {\text{epoch} = $1:N$}	
		{
			$\{(\Vec{x}_i,y_i)\}_{i=1}^{b}$ \text{sampled from data loader.} \\
			\For {$i \in \{1,...,b\}$}	
			{
				$\tilde{\Vec{x}}_i = \mathrm{aug_1}({\Vec{x}}_i)$, 
				$\tilde{\Vec{x}}'_i = \mathrm{aug_2}({\Vec{x}}_i)$ \\
				$\Vec{q}_i = E_q(\tilde{\Vec{x}}_i)$, 
				$\Vec{k}_i = E_k(\tilde{\Vec{x}}'_i)$ \\
				$\Vec{p}_i = h({\Vec{q}}_i)$ \\	
				$\Vec{s}_i=\{s_i^k\}_{k=1}^K, s_i^k = \frac{\exp(\Vec{q_i}\cdot \Vec{c}_k/\tau)}{\sum_{k=1}^{K} \exp(\Vec{q}_i\cdot \Vec{c}_k/\tau)}$ \\
				$\Vec{l}_i = (\Vec{p}_i + \Vec{s}_i)/2$ \\
				
				\uIf{epoch $ > N_s$ \text{and} $\max_k {l_i^j}>T$}
				{$\hat{y}_i=\argmax_j {\Vec{l}^j_i}$}
				\Else{$\hat{y}_i=\Vec{y}_i$}
				
				$\mathcal{L}^i_{ce} = - \log(\Vec{p}_i[\hat{y}_i])$ \\
				
				$\mathcal{L}^i_\mathrm{pro}  =  - \text{log}\frac{\text{exp}(\Vec{q}_i \cdot \Vec{c}_{\hat{y}_i} / \tau)}{\sum_{j=1}^{K} \text{exp}(\Vec{q}_i\cdot\Vec{c}_j/\tau)}$\\ 
				
				$\mathcal{L}^i_{lgc} = \frac{-1}{|\mathcal{P}({i})|} \log  \frac{ \sum\limits_{ X } \exp \left( \frac{\Vec{q}_i \cdot \Vec{k}^{+}} {\tau} \right)} { \sum\limits_{\mathclap{\Vec{k}^+ \in \mathcal{P}(i)}{}} \exp \left( \frac{\Vec{q}_i \cdot \Vec{k}^+} {\tau} \right) + \sum\limits_{\mathclap{\Vec{k}^- \in \mathcal{N}(i)}{}} \exp \left( \frac{\Vec{q}_i \cdot \Vec{k}^-} {\tau} \right)}$

				$\Vec{c}_{\hat{y}_i} = m\Vec{c}_{\hat{y}_i} + (1 - m)\Vec{q}_i$ \\
				$\mathcal{L}^i = \mathcal{L}^i_{ce} + \lambda_{pro}\mathcal{L}^i_{pro} + \lambda_{lgc}\mathcal{L}^i_{lgc}$ \\
				update networks $E_q,h$ to minimize $\mathcal{L}^i$\\
				update networks $E_q,h$ to minimize $\mathcal{L}^i$\\
				update momentum encoder $E_k$.
			}
		}
	}
	\caption{\small PNL algorithm.}
	\label{alg:pnl}
\end{algorithm}	 

\section{Compare with SYSU30K}
As show in Table~\ref{tab:lupnl_sysu}, we compare the pre-trained models between LUPerson-NL and SYSU30K. We pre-train PNL on both LUPerson-NL and SYSU30K for 40 epochs with same experiment settings for a fair comparison. The performance of LUPerson-NL pre-training is much better than SYSU30K pre-training, showing the superiority of our LUPerson-NL, and also suggesting that a large number of images with limited diversity does not bring more representative representation, but large-scale images with diversity does.

\begin{table}[t]
	\setlength{\tabcolsep}{5.5mm}
	\centering
	\begin{tabular}{l|c|c}
		\shline
		Dataset & LUPerson-NL & SYSU30K \\
		\hline
		MSMT17 & 66.1/84.6 & 55.2/76.7 \\
		\shline
	\end{tabular}
	\caption{Comparison of applying our PNL on both LUPerson-NL and SYSU30K.}
	\label{tab:lupnl_sysu}
\end{table}

\section{Hyper Parameter Analysis}
In our PNL, there are two key hyper-parameters: temperature factor $\tau$ and the threshold for correction $T$. Here, we provide the analysis of these two parameters.

\subsection{Temperature Factor $\tau$}

Table~\ref{tab:tau} shows the performance comparison with different $\tau$ values with a fixed label correction threshold $T=0.8$. As we can see, the setting $\tau=0.1$ achieves the best results on both DukeMTMC and MSMT17, while $\tau=0.07$ achieves the second best. 
When we use a larger $\tau$, the performance drops rapidly. It may be because Re-ID is a more fine-grained task, larger $\tau$ will cause smaller inter-class variations and make positive samples too close to negative samples. In all the experiments, we set $\tau=0.1$.

\subsection{Threshold $T$}

Table~\ref{tab:th} shows the results with different label correction threshold values $T$. As we can see, the performance is relatively stable to different $T$ values varying in a large range of $0.6\sim0.8$. However, if $T$ is too small or too large, the performance drops rapidly. For the former case, labels are easier to be modified, which may cause wrong rectifications, while for the latter case, the label noises become harder to be corrected, which also has consistently negative effects on the performance. In all the experiments, we set $T=0.8$.

\begin{table}[t]
	\setlength{\tabcolsep}{2.5mm}
	\begin{tabular}{l|c|c|c|c}
		\shline
		\multirow{2}{*}{$\tau$} & \multicolumn{2}{c|}{DukeMTMC} & \multicolumn{2}{c}{MSMT17} \\ 
		\cline{2-5} & 40\% & 100\% & 40\% & 100\% \\ 
		\hline 
		0.05 & 76.1/87.2 & 83.5/91.4 & 49.8/73.3 & 65.4/83.8 \\
		0.07 & \underline{76.4}/\textbf{87.5} & \underline{83.6/91.4} & \underline{50.7/74.1} & \underline{67.2/85.3} \\
		0.1  & \textbf{77.0}/\underline{87.3} & \textbf{84.3}/\textbf{92.0} & \textbf{51.9/74.9} & \textbf{68.0/86.0} \\
		0.2  & 75.8/86.8 & 83.4/91.0 & 50.1/73.7 & 66.4/85.0 \\
		0.3  & 74.7/86.2 & 82.7/90.6 & 48.6/72.0 & 65.5/83.7 \\
		\shline
	\end{tabular}
	\caption{Performances under different $\tau$ values on DukeMTMC and MSMT17 with data percentages 40\% and 100\% under the small scale setting. The threshold is set as $T=0.8$. The best scores are in bold and the second ones are underlined.}
	\label{tab:tau}
\end{table}

\begin{table}[t]
	\setlength{\tabcolsep}{2.6mm}
	\begin{tabular}{l|c|c|c|c}
		\shline
		\multirow{2}{*}{$T$} & \multicolumn{2}{c|}{DukeMTMC} & \multicolumn{2}{c}{MSMT17} \\ 
		\cline{2-5} & 40\% & 100\% & 40\% & 100\% \\ 
		\hline 
		0.5 & 76.1/86.5 & 83.3/91.0 & 51.1/74.6 & 67.5/85.2 \\
		0.6 & \textbf{77.1/87.7} & \underline{84.1}/91.6 & \textbf{52.3}/\underline{75.5} & \underline{68.1}/85.7 \\
		0.7 & 77.0/\underline{87.5} & 84.0/\underline{91.8} & \underline{51.9}/\textbf{75.6} & \textbf{68.2}/\underline{85.8} \\
		0.8 & \underline{77.0}/87.3 & \textbf{84.3/92.0} & 51.9/75.0 & 68.0/\textbf{86.0} \\
		0.9 & 75.7/86.4 & 83.0/90.8 & 50.9/74.3 & 67.2/85.0 \\
		\shline
	\end{tabular}
	\caption{Performances under different $T$ values on DukeMTMC and MSMT17 with data percentages 40\% and 100\% under the small scale setting. The temperature factor is set as $\tau=0.1$. The best scores are in bold and  the second ones are underlined.}
	\label{tab:th}
\end{table}

\section{Results for stronger backbones}
We train our PNL using two stronger backbones ResNet101 and ResNet152, and report the results in Table~\ref{tab:r101}. As we can see, the stronger ResNet bring more superior performances. These results also outperform the scores reported in Table \textcolor{red}{8} of our main submission. Most importantly, we are the \textbf{FIRST} to obtain a \emph{mAP} score on MSMT17 that is larger than $70$ without any post-processing for convolutional network.

\begin{table}[t]
	\setlength{\tabcolsep}{1.2mm}
	\centering
	\begin{tabular}{l|c|c|c|c}
		\shline
		Arch & CUHK03 & Market1501 & DukeMTMC & MSMT17 \\
		\hline
		R50  & 80.4/80.9 & 91.9/96.6 & 84.3/92.0 & 68.0/86.0 \\
		R101 & 80.5/81.2 & 92.5/\textbf{96.9} & 85.5/\textbf{92.8} & 70.8/87.1 \\
		R152 & \textbf{80.6/81.2} & \textbf{92.7}/96.8 & \textbf{85.6}/92.4 & \textbf{71.6/87.5} \\
		\shline
	\end{tabular}
	\caption{Results with different ResNet backbones. R50, R101 and R152 stand for ResNet50, ResNet101 and ResNet152 respectively.}
	\label{tab:r101}
\end{table}

\section{Pre-training data scales}

We study the impact of pre-training data scale. Specifically, we involve various percentages ($10\%$,$30\%$,$100\%$ pseudo based) of LUPerson-NL into pre-training and then evaluate the finetuing performance on the target datasets. As shown in Table \ref{tab:data-scale}, the learned representation is much stronger with the increase of the pre-training data scale, indicating the necessity of building large-scale dataset, and our LUPerson-NL is very important.

\begin{table}[t]
	\setlength{\tabcolsep}{3.4mm}
	\centering
	\begin{tabular}{l|c|c|c}
		\shline
		Scale & 10\% & 30\% & 100\% \\
		\hline
		MSMT17 & 57.4/79.2 & 62.2/82.1 & 68.0/86.0 \\
		\shline
	\end{tabular}
	\caption{Comparison for different pre-training data scale}
	\label{tab:data-scale}
\end{table}

\section{More results for small-scale and few-shot}
To complement Table 3 in the main text, we provide more detailed results under ``small scale'' and ``few shot'' settings in \Tref{tab:ss-fw}. As we can see, our weakly pre-trained models are consistently better than other pre-trained models. Our advantage is much larger with less training data, suggesting the potential practical value of our pre-trained models for real-world person ReID applications.

\begin{table*}[h]
	\setlength{\tabcolsep}{1.8mm}
	\small
	\begin{subtable}[h]{1.0\textwidth}
		\centering
		\begin{tabular}{l|cccccccccc}
			\shline
			scale & $10\%$ & $20\%$ & $30\%$ & $40\%$ & $50\%$ & $60\%$ & $70\%$ & $80\%$ & $90\%$ & $100\%$ \\
			\hline
			\#id & 75 & 150 & 225 & 300 & 375 & 450 & 525 & 600 & 675 & 751 \\ 
			\#images & 1,170 & 2,643 & 3,962 & 5,226 & 6,408 & 7,814 & 9,120 & 11,417 & 11,727 & 12,936 \\
			\hline
			IN sup.   & 53.1/76.9 & 67.7/86.8 & 75.2/90.8 & 79.1/92.5 & 81.5/93.5 & 81.5/93.5 & 84.8/94.5 & 85.9/95.2 & 86.9/95.2 & 87.5/95.1 \\ 
			IN unsup. & 58.4/81.7 & 70.2/89.1 & 76.6/91.9 & 80.0/93.0 & 82.0/94.1 & 83.7/94.3 & 85.4/94.5 & 86.4/95.0 & 87.4/95.5 & 88.2/95.3 \\ 
			LUP unsup.& 64.6/85.5 & 76.9/92.1 & 81.9/93.7 & 84.1/94.4 & 85.8/94.9 & 87.8/95.8 & 88.8/95.9 & 89.8/96.2 & 90.5/96.4 & 91.0/96.4 \\ 
			\hline
			LUPws wsp.& \textbf{72.4/88.8} & \textbf{81.7/93.2} & \textbf{85.2/94.2} & \textbf{87.3/95.1} & \textbf{88.3/95.5} & \textbf{89.6/96.0} & \textbf{90.1/96.2} & \textbf{90.9/96.4} & \textbf{91.3/96.4} & \textbf{91.9/96.6} \\
			\shline
		\end{tabular}
		\vspace{-0.12cm}
		\caption{Market1501 small-scale}
		\label{tab:ss-market}
	\end{subtable}
	
	\begin{subtable}[h]{1.0\textwidth}
		\centering
		\begin{tabular}{l|cccccccccc}
			\shline
			scale & $10\%$ & $20\%$ & $30\%$ & $40\%$ & $50\%$ & $60\%$ & $70\%$ & $80\%$ & $90\%$ & $100\%$ \\
			\hline
			\#id & 751 & 751 & 751 & 751 & 751 & 751 & 751 & 751 & 751 & 751 \\ 
			\#images & 1,293 & 2,587 & 3,880 & 5,174 & 6,468 & 7,758 & 9,055 & 10,348 & 11,642 & 12,936 \\
			\hline
			IN sup.   & 21.1/41.8 & 53.2/75.1 & 68.1/87.6 & 75.4/90.4 & 80.2/92.8 & 83.0/93.6 & 84.2/94.0 & 86.3/94.7 & 86.7/94.6 & 87.5/95.1 \\ 
			IN unsup. & 18.6/36.1 & 56.5/77.5 & 69.3/87.8 & 78.8/88.3 & 78.3/90.9 & 81.7/93.3 & 84.4/94.1 & 86.4/95.0 & 87.1/95.2 & 88.2/95.3 \\ 
			LUP unsup.& 26.4/47.5 & 63.5/83.0 & 78.3/92.1 & 80.3/92.7 & 84.2/93.9 & 86.7/94.7 & 88.4/95.5 & 89.8/96.0 & 90.4/96.3 & 91.0/96.4 \\ 
			\hline
			LUPws wsp.& \textbf{42.0/61.6} & \textbf{75.7/89.1} & \textbf{83.7/94.0} & \textbf{86.0/94.3} & \textbf{88.1/95.2} & \textbf{89.8/95.8} & \textbf{90.5/96.3} & \textbf{91.2/96.4} & \textbf{91.6/96.4} & \textbf{91.9/96.6} \\
			\shline
		\end{tabular}
		\vspace{-0.12cm}
		\caption{Market1501 few-shot}
		\label{tab:fs-market}
	\end{subtable}
	
	\begin{subtable}[h]{1.0\textwidth}
		\centering
		\begin{tabular}{l|cccccccccc}
			\shline
			scale & $10\%$ & $20\%$ & $30\%$ & $40\%$ & $50\%$ & $60\%$ & $70\%$ & $80\%$ & $90\%$ & $100\%$ \\
			\hline
			\#id & 70 & 140 & 210 & 280 & 351 & 421 & 491 & 561 & 631 & 702 \\ 
			\#images & 1,670 & 3,192 & 5,530 & 6,924 & 8,723 & 10,197 & 11,939 & 13,500 & 15,111 & 16,522 \\
			\hline
			IN sup.   & 45.1/65.3 & 58.3/75.4 & 64.7/80.2 & 68.5/83.0 & 71.8/84.6 & 74.1/85.6 & 75.5/86.8 & 76.8/87.3 & 78.0/88.3 & 79.4/89.0 \\ 
			IN unsup. & 48.1/66.9 & 60.6/76.6 & 65.8/80.2 & 69.5/82.9 & 72.5/84.4 & 75.0/86.2 & 76.3/86.9 & 77.4/87.3 & 78.5/88.7 & 79.5/89.1 \\
			LUP unsup.& 53.5/72.0 & 65.0/78.9 & 69.4/81.9 & 72.8/84.7 & 75.6/86.7 & 77.6/87.1 & 78.9/88.2 & 80.2/89.2 & 81.1/90.0 & 82.1/91.0 \\ 
			\hline
			LUPws wsp.& \textbf{60.6/75.8} & \textbf{70.5/83.3} & \textbf{74.5/86.3} & \textbf{77.0/87.3} & \textbf{78.8/88.3} & \textbf{80.5/89.2} & \textbf{81.6/89.5} & \textbf{82.9/90.6} & \textbf{83.3/91.2} & \textbf{84.3/92.0} \\
			\shline
		\end{tabular}
		\vspace{-0.12cm}
		\caption{DukeMTMC small-scale}
		\label{tab:ss-duke}
	\end{subtable}
	
	\begin{subtable}[h]{1.0\textwidth}
		\centering
		\begin{tabular}{l|cccccccccc}
			\shline
			scale & $10\%$ & $20\%$ & $30\%$ & $40\%$ & $50\%$ & $60\%$ & $70\%$ & $80\%$ & $90\%$ & $100\%$ \\
			\hline
			\#id & 702 & 702 & 702 & 702 & 702 & 702 & 702 & 702 & 702 & 702 \\
			\#images & 1,679 & 3,321 & 4,938 & 6,599 & 8,278 & 9,923 & 11,564 & 13,201 & 14,860 & 16,522 \\
			\hline
			IN sup.   & 31.5/47.1 & 56.2/72.1 & 65.4/79.8 & 71.0/83.9 & 73.9/85.7 & 75.8/86.6 & 77.2/87.8 & 78.3/88.6 & 79.1/88.8 & 79.4/89.0 \\
			IN unsup. & 32.4/48.0 & 56.4/72.2 & 65.3/80.2 & 70.2/83.4 & 73.7/85.1 & 75.8/86.7 & 77.7/88.2 & 78.7/88.7 & 79.4/89.0 & 79.5/89.1 \\
			LUP unsup.& 35.8/50.2 & 61.0/74.9 & 72.3/83.8 & 75.2/86.8 & 77.7/87.4 & 79.4/88.4 & 80.8/89.2 & 81.7/90.3 & 82.0/90.6 & 82.1/91.0 \\
			\hline
			LUPws wsp.& \textbf{52.2/64.1} & \textbf{71.7/82.5} & \textbf{77.7/87.9} & \textbf{79.3/88.1} & \textbf{81.1/89.6} & \textbf{82.3/90.2} & \textbf{83.2/91.1} & \textbf{84.0/91.6} & \textbf{84.1/91.3} & \textbf{84.3/92.0} \\
			\shline
		\end{tabular}
		\vspace{-0.12cm}
		\caption{DukeMTMC few-shot}
		\label{tab:fs-duke}
	\end{subtable}
	
	\begin{subtable}[h]{1.0\textwidth}
		\centering
		\begin{tabular}{l|cccccccccc}
			\shline
			scale & $10\%$ & $20\%$ & $30\%$ & $40\%$ & $50\%$ & $60\%$ & $70\%$ & $80\%$ & $90\%$ & $100\%$ \\
			\hline
			\#id & 104 & 208 & 312 & 416 & 520 & 624 & 728 & 832 & 936 & 1,041 \\ 
			\#images & 3,659 & 6,471 & 9,787 & 13,006 & 15,917 & 19,672 & 22,680 & 26,335 & 29,529 & 32,621 \\
			\hline
			IN sup.   & 23.2/50.2 & 34.6/\textbf{64.0} & 41.9/70.8 & 46.7/74.5 & 50.3/76.9 & 53.9/79.4 & 56.9/81.2 & 59.6/82.4 & 61.9/84.2 & 63.7/85.1 \\
			IN unsup. & 22.6/48.8 & 32.7/60.9 & 40.4/68.7 & 45.1/72.2 & 49.0/75.0 & 52.7/78.0 & 55.7/79.9 & 58.6/82.0 & 60.9/83.0 & 62.7/84.3 \\
			LUP unsup.& 25.5/51.1 & 36.0/62.6 & 44.6/\textbf{71.4} & 49.2/74.9 & 53.0/\textbf{77.7} & 56.4/79.7 & 59.5/81.8 & 61.9/\textbf{83.6} & 63.7/\textbf{85.0} & 65.7/85.5 \\
			\hline
			LUPws wsp.& \textbf{28.2/51.1} & \textbf{39.6}/63.7 & \textbf{47.7}/71.2 & \textbf{51.9/74.9} & \textbf{55.5}/77.2 & \textbf{59.1/80.1} & \textbf{61.6/81.8} & \textbf{64.2}/83.3 & \textbf{66.1}/84.8 & \textbf{68.0/86.0} \\
			\shline
		\end{tabular}
		\vspace{-0.12cm}
		\caption{MSMT17 small-scale}
		\label{tab:ss-msmt}
	\end{subtable}
	
	\begin{subtable}[h]{1.0\textwidth}
		\centering
		\begin{tabular}{l|cccccccccc}
			\shline
			scale & $10\%$ & $20\%$ & $30\%$ & $40\%$ & $50\%$ & $60\%$ & $70\%$ & $80\%$ & $90\%$ & $100\%$ \\
			\hline
			\#id & 1,041 & 1,041 & 1,041 & 1,041 & 1,041 & 1,041 & 1,041 & 1,041 & 1,041 & 1,041 \\
			\#images & 3,262 & 6,524 & 9,786 & 13,048 & 16,310 & 19,572 & 22,834 & 26,096 & 29,358 & 32,621 \\
			\hline
			IN sup.   & 14.7/34.1 & 35.6/61.4 & 44.5/71.1 & 52.0/76.9 & 56.2/79.5 & 58.8/81.7 & 60.9/82.8 & 62.5/84.2 & 63.4/84.5 & 63.7/85.1 \\
			IN unsup. & 13.2/29.2 & 33.5/58.6 & 41.4/67.1 & 47.7/72.7 & 53.3/77.6 & 56.5/79.6 & 59.1/81.5 & 60.9/82.3 & 62.4/83.8 & 62.7/84.3 \\
			LUP unsup.& 17.0/36.0 & 37.4/61.4 & 49.0/73.6 & 53.9/78.5 & 57.4/80.5 & 60.0/82.1 & 62.9/83.5 & 64.2/84.5 & 65.0/85.1 & 65.7/85.5 \\
			\hline
			LUPws wsp.& \textbf{24.5/42.7} & \textbf{45.6/67.2} & \textbf{53.2/74.4} & \textbf{58.6/78.8} & \textbf{62.2/81.0} & \textbf{64.1/82.6} & \textbf{65.8/83.8} & \textbf{67.2/84.7} & \textbf{67.4/85.3} & \textbf{68.0/86.0} \\
			\shline
		\end{tabular}
		\vspace{-0.12cm}
		\caption{MSMT17 few-shot}
		\label{tab:fs-msmt}
	\end{subtable}
	
	\caption{Performance for small-scale and few-shot setting with MGN method for Market1501, DukeMTMC and MSMT17.  ``IN sup." and ``IN unsup." refer to supervised and unsupervised pre-trained model on ImageNet, ``LUP unsup." refers to unsupervised pre-trained model on LUPerson, ``LUPws wsp.`` refers to our model pre-trained on LUPerson-WS using WSP. The first number is \emph{mAP} and second is \emph{cmc1}.}
	\label{tab:ss-fw}
\end{table*}

\end{document}